\documentclass{article}

\usepackage[preprint,nonatbib]{neurips_2025}

\usepackage[utf8]{inputenc} % allow utf-8 input
\usepackage[T1]{fontenc}    % use 8-bit T1 fonts
\usepackage{hyperref}       % hyperlinks
\usepackage{url}            % simple URL typesetting
\usepackage{booktabs}       % professional-quality tables
\usepackage{amsfonts}       % blackboard math symbols
\usepackage{nicefrac}       % compact symbols for 1/2, etc.
\usepackage{microtype}      % microtypography
\usepackage{xcolor}         % colors

\usepackage{makecell}
\usepackage{lipsum}
\usepackage{subcaption}
\usepackage{amsmath}
\usepackage{amssymb}
\usepackage{booktabs}
\usepackage{multirow}
\usepackage{multicol}
\usepackage{bbm}
\usepackage{lipsum}
\usepackage{colortbl}
\usepackage{amsthm}
\usepackage{graphicx}
\usepackage{wrapfig}
\usepackage[ruled, linesnumbered, noend]{algorithm2e}

\newcommand{\thinktoken}{\emph{\textless think\textgreater}}

\newcounter{mycounter} % create a new counter, called 'mycounter'

\usepackage{tcolorbox}

\newcommand{\findingbox}[1]{
    \stepcounter{mycounter} % Increment counter
    \begin{tcolorbox}[colframe=black,
                      arc=1pt,
                      boxsep=-2pt,
                      ]
        \noindent{\textbf{\textit{Finding \themycounter.}}} #1
    \end{tcolorbox}
}

\title{Thinkless: LLM Learns When to Think}

\author{
\bf Gongfan Fang \quad Xinyin Ma \quad Xinchao Wang\thanks{Corresponding author}\\
{National University of Singapore} \\
{\tt\small gongfan@u.nus.edu, maxinyin@u.nus.edu, xinchao@nus.edu.sg} \\
}

\begin{document}

\maketitle

\begin{abstract}
Reasoning Language Models, capable of extended chain-of-thought reasoning, have demonstrated remarkable performance on tasks requiring complex logical inference. However, applying elaborate reasoning for all queries often results in substantial computational inefficiencies, particularly when many problems admit straightforward solutions. This motivates an open question: Can LLMs learn when to think? To answer this, we propose Thinkless, a learnable framework that empowers an LLM to adaptively select between short-form and long-form reasoning, based on both task complexity and the model's ability. Thinkless is trained under a reinforcement learning paradigm and employs two control tokens, \emph{<short>} for concise responses and \emph{<think>} for detailed reasoning. At the core of our method is a Decoupled Group Relative Policy Optimization (DeGRPO) algorithm, which decomposes the learning objective of hybrid reasoning into two components: (1) a control token loss that governs the selection of the reasoning mode, and (2) a response loss that improves the accuracy of the generated answers. This decoupled formulation enables fine-grained control over the contributions of each objective, stabilizing training and effectively preventing collapse observed in vanilla GRPO. Empirically, on several benchmarks such as Minerva Algebra, MATH-500, and GSM8K, Thinkless is able to reduce the usage of long-chain thinking by 50\% - 90\%, significantly improving the efficiency of Reasoning Language Models. The code is available at \url{https://github.com/VainF/Thinkless}
\end{abstract}

\section{Introduction}

Reasoning Language Models have exhibited notable effectiveness in solving complex tasks that involve multi-hop reasoning and logic-intensive inference. Their capabilities span a range of domains, such as mathematical problem solving~\cite{jaech2024openai,guo2025deepseek,shao2024deepseekmath} and agentic assistants~\cite{wu2025agentic,chen2025research}. A primary factor underlying this success is their ability to perform chain-of-thought reasoning~\cite{wei2022chain}, wherein intermediate steps are explicitly generated before arriving at a final answer. While this approach is effective for solving challenging problems, applying it uniformly to all queries, regardless of their complexity or the model’s capability can be inefficient. In particular, for questions with straightforward solutions, invoking extended reasoning results in redundant token generation, increased memory footprint, and substantially higher computational cost~\cite{feng2025efficient,chen2024not}.

In response to these inefficiencies, a growing body of research has sought to enhance the inference efficiency of reasoning models~\cite{feng2025efficient,aggarwal2025l1,ma2025cot,luo2025o1,cuadron2025danger,hashemi2025dna,srivastava2025towards}. A prominent direction in this space explores \emph{hybrid reasoning}~\cite{bercovich2025llamanemotronefficientreasoningmodels,qwen3,claude2025}, wherein models dynamically switch between reasoning and non-reasoning modes~\cite{yang2024qwen2,claude2025}. 
Despite promising results, a central challenge persists: determining when a model should engage in elaborate reasoning. Many existing approaches address this by incorporating manually designed heuristics, such as fixed computational budgets~\cite{aggarwal2025l1} or prompt-level control signals like ``reasoning on/off''~\cite{bercovich2025llamanemotronefficientreasoningmodels,qwen3}. However, these strategies inherently rely on human prior knowledge and may yield suboptimal or inappropriate control decisions. This underscores a fundamental open question: Can an LLM \emph{learn} to decide when to think, guided by the complexity of the task and its own capability?

Motivated by this, we explore the fundamental form of hybrid reasoning, where the model is tasked with autonomously deciding whether to generate a short-form or long-form response based on the input query. This decision is guided by three core factors: (1) the complexity of the query, as simpler questions generally merit concise responses, while more intricate ones may necessitate extended reasoning; (2) the capability of the model, since more powerful models are better positioned to employ short reasoning without sacrificing accuracy, whereas less capable models may benefit from longer responses to preserve performance; and (3) the user's tolerance for the trade-off between efficiency and accuracy, which determines the acceptable level of performance degradation when opting for shorter reasoning. Naturally, reinforcement learning~\cite{shao2024deepseekmath,guo2025deepseek,zeng2025simplerl} offers a framework to unify these factors, as it allows the model to learn from interactions that reflect both environmental feedback and user-defined preferences. Through iterative exploration and reward-driven updates, the model progressively acquires the ability to make autonomous, context-aware decisions about its reasoning strategy, balancing accuracy and efficiency in a dynamic and data-driven manner.

\begin{figure}
    \centering
    \includegraphics[width=\linewidth]{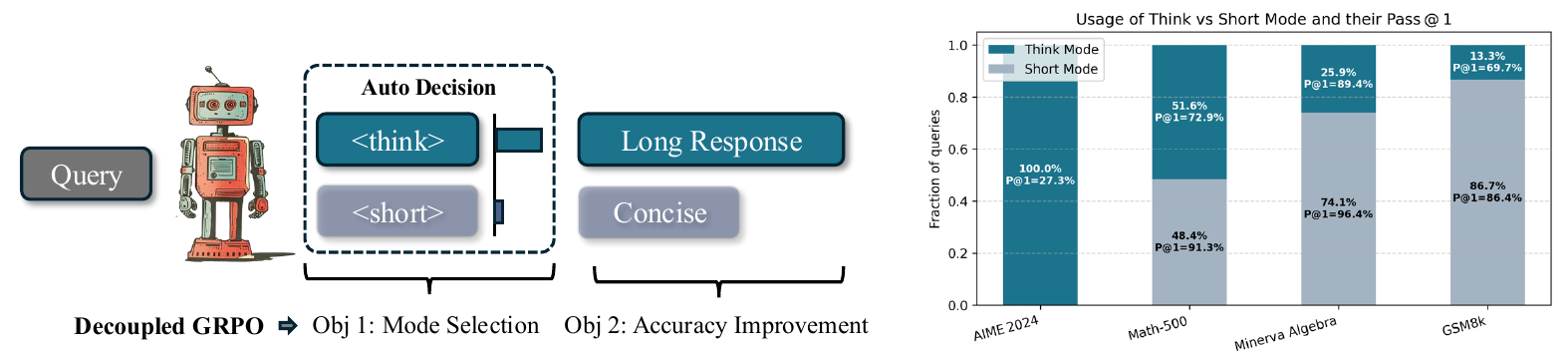}
    \caption{Thinkless learns a hybrid LLM capable of adaptively selecting between thinking and non-thinking inference modes, directed by two special tokens, \emph{<think>} and \emph{<short>}. At the core of our method is a Decoupled Group Relative Policy Optimization, which decomposes and balances the mode selection on the control token and accuracy improvement on the response tokens.}
    \label{fig:intro}
\end{figure}

Building on these insights, we propose Thinkless, a reinforcement learning framework designed to train a hybrid reasoning model capable of selecting between short-form and long-form responses. As illustrated in Figure 3, Thinkless employs two control tokens, \emph{<think>} and \emph{<short>}, which are generated as the first token in the model’s output to signal the intended inference style. The training comprises two stages: a supervised warm-up phase followed by a reinforcement learning phase.

\noindent\textbf{Distillation for Warm-up.} In the warm-up phase, the model aligns its response style with the designated control tokens via a distillation process. Specifically, it learns to imitate the behavior of two expert models: a reasoning model and a standard instruction-following model, each conditioned on a specific control token (\emph{<think>} or \emph{<short>}). Additionally, the model is trained on paired long-form and short-form responses for each query, ensuring it can generate both styles with comparable likelihood. This initialization establishes a clear and robust mapping between control tokens and response formats, providing diverse outputs for subsequent reinforcement learning.

\noindent\textbf{Reinforcement Learning with Decoupled GRPO.} In the reinforcement learning phase, the model is optimized to select the appropriate inference mode based on performance feedback. A natural starting point for this task is the vanilla Group Relative Policy Optimization (GRPO)~\cite{guo2025deepseek} framework. However, when applied to hybrid reasoning, vanilla GRPO treats all tokens, including the control token and the response tokens uniformly. This introduces a critical imbalance: since the response part often spans hundreds to thousands of tokens and the length of long \& short responses varies significantly, the single control token may receive weak and biased gradient signals, ultimately leading to mode collapse at the early stages of training. To this end, we propose a tailored method for hybrid reasoning, termed Decoupled Group Relative Policy Optimization (DeGRPO). As illustrated in Figure~\ref{fig:intro}, DeGRPO explicitly separates the hybrid reasoning objective into two components:  (1) Mode Selection, which governs how quickly the policy adapts based on the model’s current accuracy; and (2) Accuracy Improvement, which refines the response content to improve answer correctness under the selected reasoning mode. These two components are inherently interdependent, and effective training requires carefully balancing the learning signals for the control token and the response tokens. For instance, if the mode selection policy adapts too aggressively in favor of long-form reasoning, the model may ignore short-form responses, resulting in insufficient exploration of the potential of short responses. To this end, DeGRPO assigns distinct weights to the control token and response tokens, promoting a more stable and balanced training dynamic. This design not only mitigates mode collapse but also enables the model to learn both accurate outputs and context-sensitive reasoning strategies. After reinforcement learning, the model learns to accurately identify simple queries and respond using the more efficient non-thinking mode. For instance, on benchmarks such as MATH-500, Minerva Algebra and GSM8K dataset, Thinkless reduces the usage of long-form reasoning by 50\% - 90\%. And on much more challenging tasks like AIME, the model naturally adopts a higher proportion of long-form reasoning. 

In conclusion, this work demonstrates that a Reasoning Language Model can learn when to engage in reasoning before generating a response, guided by the proposed Decoupled GRPO method. This adaptive decision-making substantially reduces inference cost while preserving task performance.

\section{Related Works}

\paragraph{Efficient Reasoning Models.}
Reasoning models generate intermediate steps in a chain-of-thought process before producing a final answer~\cite{wei2022chain,jaech2024openai}. This paradigm offers significant advantages for tasks involving complex computations, logical deduction, and multi-step reasoning. However, excessively long reasoning chains can lead to substantial computational overhead~\cite{chen2024not,cuadron2025danger,feng2025efficient}. To mitigate this, recent research has explored strategies to enhance the efficiency of reasoning models without sacrificing accuracy or generalization~\cite{ma2025cot,luo2025o1,aggarwal2025l1,kang2024c3ot,han2024token}. Several techniques, such as reinforcement learning with length penalties~\cite{aggarwal2025l1,luo2025o1}, supervised fine-tuning using variable-length chain-of-thought data~\cite{ma2025cot}, and prompt-based methods~\cite{chen2024not,xu2025chain,aytes2025sketch} have been proposed to encourage concise yet effective reasoning paths. Additionally, latent reasoning techniques aim to encode reasoning steps into compact internal representations, thereby reducing token-level computation while maintaining performance~\cite{saunshi2025reasoning}. Parallel efforts in knowledge distillation~\cite{hinton2015distilling,yu2024distilling,magister2022teaching,li2025small,chen2024distilling} have facilitated the transfer of reasoning capabilities from large to smaller models, while efficient decoding strategies such as predictive decoding and self-consistency optimization have also yielded notable improvements in inference speed and resource utilization~\cite{guo2025deepseek,li2023mixed,zhu2024improving}.

\paragraph{Hybrid Reasoning.}
While prior work has largely focused on compressing reasoning paths to reduce token generation, an alternative path to efficiency is hybrid reasoning, which dynamically adapts the appropriate inference behaviour based on task complexity~\cite{claude2025}. This approach allows models to flexibly alternate between short-form responses and long-chain reasoning as needed. Hybrid reasoning can be realized either through collaborative systems involving multiple models~\cite{ong2406routellm,liao2025reward} or within a single unified model~\cite{claude2025,qwen3,bercovich2025llamanemotronefficientreasoningmodels}. In multi-model frameworks, routing mechanisms~\cite{ong2406routellm} or speculative decoding techniques~\cite{liao2025reward} are commonly employed. For example, a lightweight model may generate a preliminary answer that a larger model verifies or refines. In contrast, unified models are trained to support both reasoning modes and can switch between them via prompt-based control. Some models adopt fixed prompt formats such as “reasoning on/off” to modulate reasoning depth~\cite{bercovich2025llamanemotronefficientreasoningmodels}. However, most existing approaches depend on manually crafted heuristics to balance efficiency and performance. In this work, we explore a learning-based alternative, enabling an LLM to automatically determine its inference behavior based on inputs, without relying on manual control.

\section{Method}

The proposed Thinkless is implemented in two stages: (1) \textbf{Distillation for Warm-up}, where we fine-tune a pre-trained reasoning model to unify two reasoning styles, and (2) \textbf{Reinforcement Learning with DeGRPO}, where the model is trained under a decomposed objective to select the appropriate reasoning mode while improving the response quality.

\subsection{Distillation for Warm-up}

The first step in our framework is to construct a model $\pi_\theta$ capable of generating both short- and long-form responses. We leverage two pre-trained experts for distillation: a reasoning model $\pi_{\text{think}}$, trained to produce detailed chains of thought via step-by-step reasoning; and an instruction-following model $\pi_{\text{short}}$, optimized for generating concise answers aligned with user intent.

\begin{figure}
    \centering
    \includegraphics[width=0.96\linewidth]{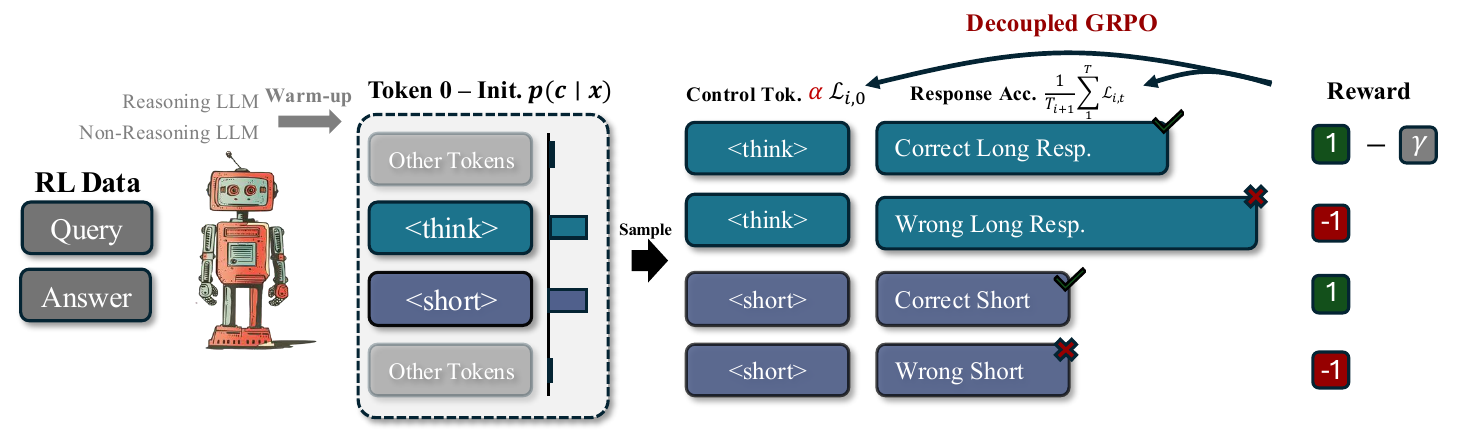}
    \caption{ThinkLess trains a hybrid model that adaptively selects reasoning modes based on task complexity and model capacity. The process begins with distillation, enabling the model to follow control tokens (\emph{<think>} or \emph{<short>}) for guided reasoning. This is followed by reinforcement learning using Decoupled GRPO, which separates training into two objectives: optimizing the control token for effective mode selection and refining the response to improve answer accuracy.}
    \label{fig:framework}
\end{figure}

Given a prompt corpus $\mathcal{X} = \{x_i\}_{i=1}^N$, we use these models to generate a synthetic paired dataset:
\[
\mathcal{D}_{\text{distill}} = \left\{ \left(x_i, \texttt{<think>} \ a_i^{\text{think}}, \texttt{<short>} \ a_i^{\text{short}} \right) \right\}_{i=1}^N,
\]
where $a_i^{\text{think}} = \pi_{\text{long}}(x_i)$ and $a_i^{\text{short}} = \pi_{\text{short}}(x_i)$. Each response is prefixed with a control token $c \in \mathcal{C} = \{\texttt{<short>}, \texttt{<think>}\}$ that conditions the model on the intended reasoning style. We then fine-tune the target reasoning model $\pi_\theta$ on this dataset via supervised fine-tuning (SFT). The objective is to learn a multi-style response distribution conditioned on the control token. This distillation phase ensures that the model is capable of generating both types of responses with high fidelity. Moreover, the paired construction of $\mathcal{D}_{\text{distill}}$ ensures that, the model's response distribution will be balanced. This helps the follow-up RL process to explore different solutions. 

\subsection{Learning When to Think via Decoupled GRPO}

After the distillation phase, the model can produce both long‑ and short‑form answers. what it still lacks is a mechanism for \emph{deciding} which reasoning mode suits a particular input~\(x\).
To supply this capability we frame mode selection as a reinforcement‑learning problem and optimize a policy \(\pi_\theta(c,a\mid x)=\pi_\theta(c\mid x)\,\pi_\theta(a\mid x,c)\), where the first token $c \in \mathcal{C}=\{\texttt{<short>}, \texttt{<think>}\}$ serves as a \emph{control token} that determines the reasoning mode, and the subsequent tokens $(a_{i,1}, \dots, a_{i,T_i})$ constitute the generated response. For notational convenience, we denote the entire sequence of the $i$-th sample of length $T_i+1$ as $a_i = (a_{i,0}, \dots, a_{i,T_i})$, where $a_{i,0} \in \mathcal{C}$ is the control token.

\noindent\textbf{Reward Design.}
Let \( y^* \) denote the ground-truth answer corresponding to the input \( x \). We consider a minimally designed reward function \( r(a, y^*, c) \), which assigns different reward values as follows:
\[
r(a, y^*, c) =
\begin{cases}
1.0, & \text{if } c = \emph{\text{<short>}} \text{ and } \text{Extract-Answer}(a) = y^*, \\
1.0 - \gamma, & \text{if } c = \emph{\text{<think>}} \text{ and } \text{Extract-Answer}(a) = y^*, \\
-1.0, & \text{if } \text{Extract-Answer}(a) \neq y^*, \\
\end{cases}
\]
where the $1>\gamma>0$ introduces preference to the short correct answer over long responses. 

\noindent\textbf{Decoupled Policy Optimization.}
Based on the simple reward function, we adopt a GRPO-based framework~\cite{shao2024deepseekmath,liu2025understanding} for training.
Let \(\{a_i\}_{i=1}^{G}\) denote a mini‑batch of trajectories sampled from the current policy \(\pi_{\theta_{\mathrm{old}}}\). The objective is defined as:
\begin{equation}
\label{eq:objective}
\small
\mathcal{J}_{\mathrm{GRPO}}(\theta)=
\mathbb{E}_{x,a_i}\Biggl[
\frac{1}{G}\sum_{i=1}^{G}
\Bigl(
\tfrac{1}{T_i+1}\sum_{t=0}^{T_i}\mathcal{L}_{i,t}(\theta)
-\beta\,\mathbb{D}_{\mathrm{KL}}\!\bigl[\pi_\theta(\cdot\mid x)\,\|\,\pi_{\mathrm{ref}}(\cdot\mid x)\bigr]
\Bigr)
\Biggr],
\end{equation}
where \(\mathcal{L}_{i,t}(\theta)\) denotes the token-level surrogate loss formally given by:
\begin{equation}
\begin{aligned}
\mathcal{L}_{i,t}(\theta) =
\min \left( \frac{\pi_\theta(a_{i,t} \mid x, a_{i,<t})}{\pi_{\theta_{\mathrm{old}}}(a_{i,t} \mid x, a_{i,<t})} \hat{A}_{i,t}, \right. \left. \text{clip} \left( \frac{\pi_\theta(a_{i,t} \mid x, a_{i,<t})}{\pi_{\theta_{\mathrm{old}}}(a_{i,t} \mid x, a_{i,<t})}, 1 - \epsilon, 1 + \epsilon \right) \hat{A}_{i,t} \right)
\end{aligned}
\label{eq:clipped_loss}
\end{equation}
In this work, we compute the relative advantage using \(\hat{A}_{i,t} = r - \mathrm{mean}(\mathbf{r})\), following~\cite{liu2025understanding}. This choice is motivated by the observation that our training data contains questions of varying difficulty, which can introduce bias when using standard deviation normalization.

When applied to training, the objective in Equation~\ref{eq:objective} serves two purposes: (1) to learn an appropriate control token for mode selection, and (2) to improve the accuracy of the response tokens:
\begin{equation}
\label{eq:original_loss}
\frac{1}{T_i+1}\sum_{t=0}^{T_i}
\mathcal{L}_{i,t}(\theta) =
\underbrace{\frac{1}{T_i+1} \mathcal{L}_{i,0}(\theta)}_{\text{Control Token}} +
\underbrace{\frac{1}{T_i+1} \sum_{t=1}^{T_i} \mathcal{L}_{i,t}(\theta)}_{\text{Response Tokens}}.
\end{equation}
For mode selection, the response style is conditioned on the first token \(a_{i,0}\), which is trained during the preceding distillation stage. As a result, adjusting the probability of this single control token is sufficient to switch between reasoning modes. Therefore, this token controls the learning of inference mode. For response accuracy, the optimization seeks to improve the generation of the remaining tokens \(a_{i,1:T_i}\). However, the above Equation~\ref{eq:original_loss} introduces two types of imbalance during optimization:  
(1) \emph{Mode-Accuracy imbalance} - each trajectory contains only one control token but \(T_i\) response tokens, disproportionately reducing the influence of the mode selection compared to the optimization of response accuracy.
(2) \emph{Think-Short imbalance} - longer sequences (\(T_i^{\text{think}} \gg T_i^{\text{short}}\)) further suppress the gradient contribution of the control token due to the normalization factor \(1/(T_i + 1)\), causing the \emph{<think>} token to be under-optimized compared to the \emph{<short>}. As we will show in the experiment, such imbalance may lead to severe mode-collapse at the beginning of training. To mitigate these imbalances, we propose a decoupled variant of GRPO, denoted as \(\mathcal{J}_{\mathrm{DeGRPO}}\), which separately normalizes the contributions of the control and response tokens:
\begin{equation}
\label{eq:DeGRPO}
\mathcal{J}_{\mathrm{DeGRPO}}(\theta)=
\mathbb{E}_{x,a_i}\Biggl[
\frac{1}{G}\sum_{i=1}^{G}
\Bigl(
 \underbrace{\alpha\, \mathcal{L}_{i,0}(\theta)}_{\text{Control Token}} + \underbrace{\frac{1}{T_i} \sum_{t=1}^{T_i} \mathcal{L}_{i,t}(\theta)}_{\text{Response Tokens}}
- \beta\, \mathbb{D}_{\mathrm{KL}}\bigl[\pi_\theta(\cdot \mid x)\,\|\,\pi_{\mathrm{ref}}(\cdot \mid x)\bigr]
\Bigr)
\Biggr],
\end{equation}
In DeGRPO, the mode selection \(\mathcal{L}_{i,0}(\theta)\) and response accuracy improvement \(\sum_{t=1}^{T_i} \mathcal{L}_{i,t}(\theta)\) are independently normalized. A length-independent weighting coefficient \(\alpha\) is introduced to balance the optimization between mode selection and response generation. This formulation ensures that the control token receives a consistent gradient scale across both short and long sequences, thereby addressing the mode-mode and think-short imbalance, enabling more stable optimization of reasoning-mode selection. As will be shown in the experiments, an appropriately large $\alpha$ can make the mode update more efficient. In our experiment, we set $\alpha=1/1000$ for stable training. 

\paragraph{Method Summary.} In summary, our method retains the overall structure of the standard GRPO framework.
For each query, a mini-batch of samples is drawn from the current policy to estimate the token-level advantages.
To address the imbalance between mode selection and response generation, we independently normalize the advantages associated with the control token and response tokens.
This separation allows for explicit balancing of their contributions during optimization, leading to more stable and effective training.

\section{Experiments}

\subsection{Experimental Setups}

\paragraph{LLMs and Datasets.} We employ \texttt{DeepSeek-R1-Distill-Qwen-1.5B} as the base model to train a hybrid reasoning policy. To construct long-short paired responses for the distillation phase, we utilize long-form data from open-source datasets~\cite{OpenThoughts,openr1} generated by the \texttt{DeepSeek-R1-671B} model~\cite{guo2025deepseek}, which is well-suited for multi-step reasoning. The corresponding short-form answers are derived using \texttt{Qwen2.5-Math-1.5B-Instruct}~\cite{yang2024qwen2}, a compact instruction-tuned model optimized for concise mathematical responses. The hybrid model is direcytly fine-tuned on this paired dataset via supervised fine-tuning, enabling it to accommodate both long and short reasoning styles. The model is then further optimized using the Decoupled Group Relative Policy Optimization (GRPO) algorithm. For the reinforcement learning stage, we primarily use the \texttt{DeepScaleR} dataset~\cite{luo2025deepscaler}, which comprises approximately 40K labeled examples. For evaluation, we mainly focus on math datasets, including AIME~\cite{aime_1983_2024}, Minerva Algebra~\cite{hendrycksmath2021}, MATH-500~\cite{lightman2023lets} and GSM-8K~\cite{cobbe2021gsm8k}. 

\paragraph{Training Details.} 
All experiments were conducted on a single node with 4 H100 GPUs. For the warmup stage, we set the maximum context length to 16K and clip those overlong samples. We train the \texttt{DeepSeek-R1-Distill-Qwen-1.5B} for only 1 epoch. The SFT was conducted on the Megatron framework~\cite{shoeybi2019megatron}. For the reinforcement learning stage, we extend the context length to 24K. The model was trained only for 600 steps, using the AdamW optimizer with a learning rate of $1 \times 10^{-6}$, $\beta = (0.9, 0.999)$, and a weight decay of 0.01. The batch size is set to 128, with 8 responses sampled for each query, leading to 1024 data points in total. The RL experiments were implemented using the VeRL framework~\cite{sheng2024hybridflow}. More details can be found in the Appendix. 

\begin{table*}[t]
\centering
\small
\resizebox{\textwidth}{!}{
\begin{tabular}{l c c c c c c c c c}
\toprule
\multirow{2}{*}{\bf Models} & \bf \multirow{2}{*}{\bf Type} & \multicolumn{2}{c}{\textbf{AIME 2024}} & \multicolumn{2}{c}{\textbf{Minerva Algebra}} & \multicolumn{2}{c}{\textbf{Math-500}} & \multicolumn{2}{c}{\textbf{GSM8K}}\\
\cmidrule(lr){3-4} \cmidrule(lr){5-6} \cmidrule(lr){7-8}  \cmidrule(lr){9-10}
& & Pass@1 & \#Tokens (Think\%) & Pass@1 & \#Tokens (Think\%) & Pass@1 & \#Tokens (Think\%) & Pass@1 & \#Tokens (Think\%) \\

\midrule
\textbf{DeepSeek-R1-1.5B}  &  \multirow{3}{*} & 0.2800 &  18063 & 0.9577 & 3029 & 0.8608 & 5675 & 0.8347 & 1919 \\
\textbf{Q-1.5B} & {Base LLM} & 0.0200 & 1300 &  0.7771 & 933 & 0.5168 & 855 & 0.7022 & 466 \\
\textbf{QMath-1.5B} &   & 0.1133 & 1128 & 0.9184 & 586 & 0.7604 & 721 & 0.8572 & 447 \\

\midrule

\bf Merging-0.5~\cite{team2025kimi} & \multirow{6}{*}{Short CoT} & 0.1333 & 8636 & 0.9292 & 834 & 0.7740 & 1524 & 0.8332 & 601 \\
\bf Merging-0.6~\cite{team2025kimi} &  & 0.1733 & 10615 & 0.9321 & 
1091 & 0.7900 & 3000 & 0.8381 & 747 \\
\bf Merging-0.7~\cite{team2025kimi} &  & 0.1667 & 15854 & 0.9398 & 1834 & 0.8108 & 4347 & 0.8458 & 1201 \\
\bf CoT-Valve $\alpha=8$~\cite{ma2025cot} & & 0.2000 & 10692 & 0.8079 & 1903 & 0.7060 & 3723 & 0.7726 & 773 \\
\bf CoT-Valve $\alpha=6$~\cite{ma2025cot} & & 0.1933 & 17245 & 0.9468 & 2656 & 0.8024 & 5167 & 0.7970 & 1009 \\
\bf CoT-Valve $\alpha=4$~\cite{ma2025cot} & & 0.2267 & 17722 & 0.9439 & 2965 & 0.8036 & 5820 & 0.8108 & 1396 \\

\midrule

\bf Router Random & & 0.1467 & 8093 (56.00\%) & 0.9211 & 1736 (49.28\%) & 0.7608 & 3096 (47.92\%) & 0.8205 & 1086 (50.99\%)   \\

\bf Router Q-7B & & 0.1667 & 9296 (46.67\%) & 0.9250 & 795 (5.64\%) & 0.7948 & 2748 (25.00\%) & \bf 0.8587 &  563 (2.35\%) \\

\bf Thinkless & \multirow{-3}{*}{Hybrid}   & \bf 0.2733 &  7099 (100.00\%) & \bf 0.9459 & 1144 (25.88\%) & \bf 0.8184 & 2555 (51.56\%) & 0.8418 & 624 (13.31\%) \\

\bottomrule
\end{tabular}
}
\caption{Empirical results of hybrid reasoning. For hybrid algorithms, we additionally report the proportion of queries executed in the thinking mode during evaluation.}\label{tbl:main}
\end{table*}

\subsection{Empirical Results on Hybrid Reasoning}

\findingbox{The learned hybrid reasoning models effectively distinguish complex from simple queries, reducing the use of thinking by 50\%--90\%.}

Table~\ref{tbl:main} presents a comparison between our method and several existing models or techniques. The first part showcases our baseline model, \texttt{DeepSeek-R1-Distill-Qwen-1.5B}, alongside two instruction-following models designed to generate concise answers. It can be observed that the number of tokens generated by reasoning models is typically 5 to 20 times higher than that of standard models. This highlights the potential for significantly improving efficiency by appropriately selecting the reasoning mode. Notably, on challenging datasets such as AIME and MATH-500, reasoning models tend to outperform others by a large margin. However, on simpler datasets like GSM-8K, extended reasoning offers no clear advantage, and the \texttt{Qwen2.5-Math-1.5B-Instruct} model with only a 4K context length achieves better results.

The second part of the table illustrates techniques for generating shorter chains of thought. One highly effective approach is model merging~\cite{team2025kimi}, where \texttt{DeepSeek-R1-Distill-Qwen-1.5B} is interpolated in parameter space with a base model, \texttt{Qwen2.5-Math-1.5B}, to obtain a more efficient model without additional training. Another method is the CoT-Valve~\cite{ma2025cot} technique, which applies supervised fine-tuning (SFT) using LoRA. It allows for controllable reasoning length by adjusting the $\alpha$ parameter in LoRA to modulate the magnitude of parameter updates. Both methods provide mechanisms for adjusting reasoning length, such as the interpolation ratio and the LoRA $\alpha$. To show their performance, we sample outputs across a range of lengths. A notable challenge with these methods is that the optimal reasoning length varies significantly across datasets. Consequently, when a good heuristic, such as merging coefficients of 0.6, is determined on one dataset to reduce token usage, it may result in unexpected performance degradation on other benchmarks, such as AIME. 

In the final part of our study, we examine hybrid reasoning strategies, focusing on a comparison with router-based approaches. These methods employ a separate large language model (LLM) to assess query difficulty and dispatch inputs to either a reasoning model or a standard model. While effective to some extent, router models are typically independent and lack comprehensive awareness of the target reasoning models, limiting their ability to make confidence-informed decisions. For instance, on the challenging AIME dataset, where even the reasoning model achieves only 28\% accuracy, the router struggles to recognize the its difficulty. In contrast, our method jointly considers both input complexity and model capability, dynamically refining the dispatch strategy through direct interaction with real examples. As a result, it achieves efficient and adaptive reasoning without manual tuning. On the Minerva Algebra dataset, our approach activates the reasoning mode for only 25\% of the samples, reducing token usage to one-third of the original, while maintaining performance within a 1\% margin. In addition, we found that the RL will also compress the length of long responses, since the algorithm will encourage short and correct answers, producing a gradient towards a more compact response.

\subsection{Training Dynamics in RL} 
\findingbox{Policy may collapse due to imbalanced update of control tokens in Vanilla GRPO.}
\paragraph{Mode Collapse in RL.}

\begin{figure}[t]
    \centering
    % -----------------------------------------------------------
    % (a)  stand‑alone panel + caption
    % -----------------------------------------------------------
    \begin{subfigure}[b]{0.242\linewidth}
        \centering
        \includegraphics[width=\linewidth]
            {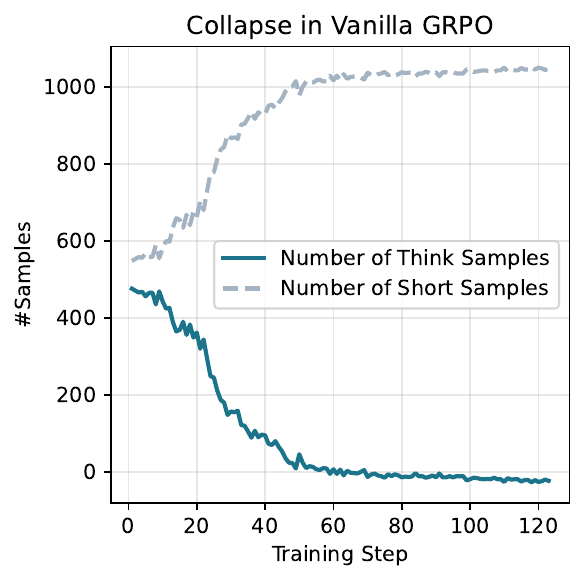}
        \caption{Vanilla GRPO}   % ← caption for the first panel
        \label{fig:collapse:a}
    \end{subfigure}
    \hfill
    \vrule
    \hfill
    % -----------------------------------------------------------
    % (b–d)  three panels grouped inside one subfigure
    %        so they share a single caption
    % -----------------------------------------------------------
    \begin{subfigure}[b]{0.745\linewidth}        % 0.24×3 = 0.72
        \centering
        \begin{subfigure}[b]{0.32\linewidth}
            \includegraphics[width=\linewidth]
                {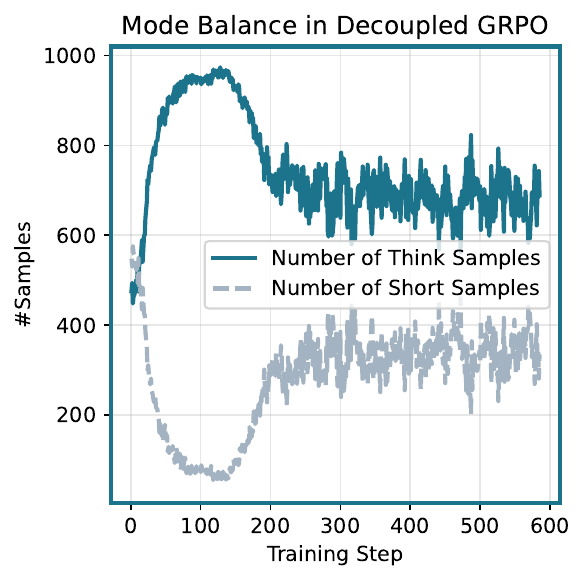}
        \end{subfigure}
        \hfill
        \begin{subfigure}[b]{0.32\linewidth}
            \includegraphics[width=\linewidth]
                {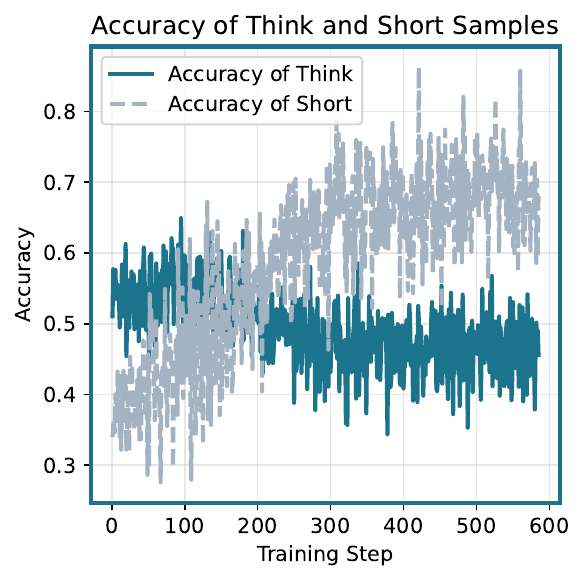}
        \end{subfigure}
        \hfill
        \begin{subfigure}[b]{0.32\linewidth}
            \includegraphics[width=\linewidth]
                {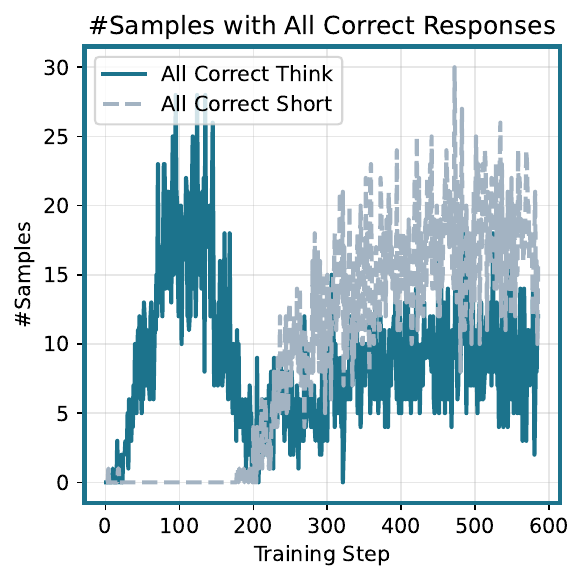}
        \end{subfigure}

        \caption{The proposed Decoupled GRPO, with a U-shape learning curve.}
        \label{fig:collapse:DeGRPO}
    \end{subfigure}

    \vspace{4mm}

    \begin{subfigure}[b]{0.48\linewidth}
        \centering
        \includegraphics[width=\linewidth]{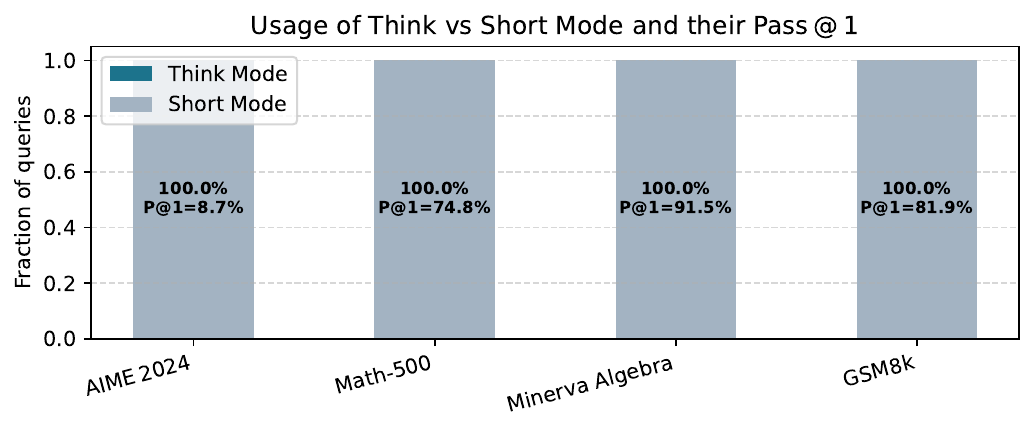}
        \caption{Collapsed Policy}
        \label{fig:collapse:c}
    \end{subfigure}
    \hfill
    \begin{subfigure}[b]{0.48\linewidth}
        \centering
        \includegraphics[width=\linewidth]{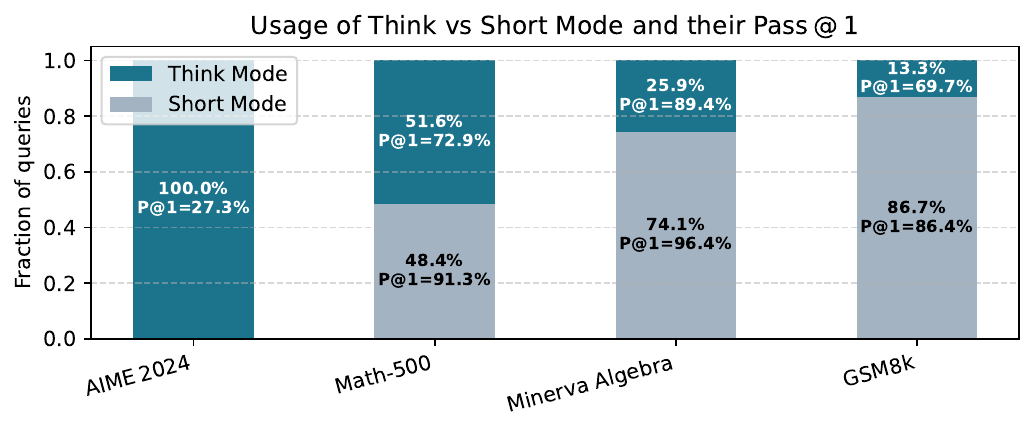}
        \caption{Learned Policy of DeGRPO}
        \label{fig:collapse:b}
    \end{subfigure}
    
    % -----------------------------------------------------------
    % Overall figure caption
    % -----------------------------------------------------------
    \caption{Policy–training comparison between vanilla GRPO and decoupled GRPO.}
    \label{fig:training_dynamics}
\end{figure}

To further analyze how the model learns a reasonable policy, we visualize the training process of RL. Figure~\ref{fig:training_dynamics} (a) illustrates the \emph{Mode Collapse} issue in standard GRPO, where the model develops an excessive preference for either long or short outputs during training. In conventional GRPO, the gradient on the control token is normalized by the total length of the response, which introduces an imbalance between long and short outputs. Specifically, long-chain samples, due to having more tokens, receive slower updates on the \texttt{<think>} token, while samples encouraging \texttt{<short>} dominate the updates. This imbalance causes the model to collapse rapidly, as shown in Figure~\ref{fig:training_dynamics} (a), the number of generated long-chain responses drops below 10 within just 120 update steps, making it difficult for the model to learn the correct policy. Furthermore, as shown in Figure~\ref{fig:training_dynamics} (c), the model fails to correctly differentiate between samples of varying difficulty, consistently opting for the short-chain reasoning mode.

\findingbox{The U-shape learning curve: The proportion of short-chain samples first drops due to low initial accuracy, then rises after accuracy improvement and mode selection take effect.}
\paragraph{The U-Shape Learning Curve.} 
To mitigate the collapse issue, we propose a Decoupled GRPO algorithm. Figure~\ref{fig:training_dynamics} (b) illustrates the impact of this decoupling on the training process. We observe a characteristic U-shaped curve in the RL process: the proportion of long-chain outputs initially increases and then gradually decreases. After properly balancing the update weights between the control token and response tokens, the model shows a preference for long-chain reasoning in the early stages of training, primarily because long chains tend to yield higher accuracy. As training progresses, we observe an improvement in the accuracy of short-chain responses. This is driven by two factors: (1) reinforcement learning enhances the generation quality, and (2) the model learns to assign simpler queries to the short-chain reasoning mode. As a result, short-chain responses receive increasingly higher rewards, encouraging the model to further explore the feasibility of short reasoning. This behavior is manifested in the rising proportion of short-chain outputs over time, with many short responses achieving perfect accuracy in the later stages of training. Additionally, we observe a decline in the accuracy of long-chain responses during the latter half of training. This phenomenon is not due to a degradation of the model’s reasoning ability but rather a shift in task allocation: more difficult queries, which cannot be solved via short reasoning, are assigned to the long-chain mode, thereby lowering its average accuracy.

\findingbox{The weight $\alpha$ of the control token governs the learning speed of mode selection.}
\paragraph{The Influence of Decoupling.} 
\begin{wrapfigure}{r}{0.35\textwidth}
    \centering
    \vspace{-6mm}
    \includegraphics[width=\linewidth]{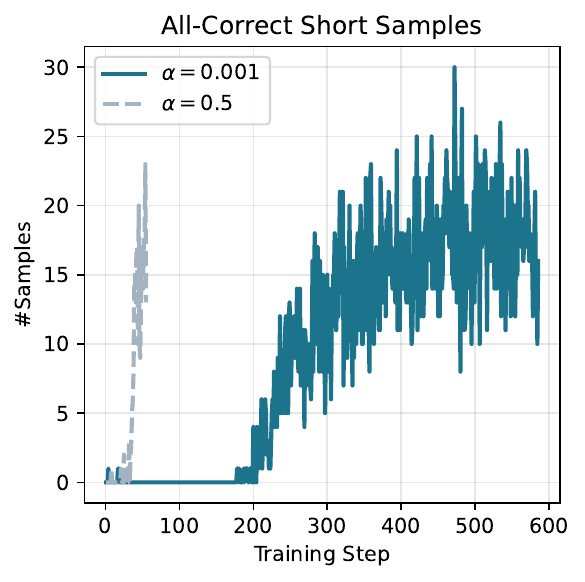}
    \caption{A large token loss coefficient $\alpha$ accelerates the shift in reasoning behavior, leading to the rapid emergence of all-correct short-mode samples.}\label{fig:decoupling}
\end{wrapfigure}
Building upon the above analysis, we further visualize the effect of decoupling on model behavior. In Figure~\ref{fig:decoupling}, we show the number of samples that are correctly answered with short responses across all sampled trajectories. We compare the results under two settings: a high control token update weight (0.5) and the weight used in our method (0.001). We observe that with a higher weight on the control token, the all-correct short samples emerge earlier in training. This is because the control token is updated more aggressively, allowing the model to focus more on learning the mode selection. However, excessively fast policy updates can be problematic. For example, some samples may initially yield low accuracy under short responses, but reinforcement learning could eventually improve their short-mode performance. If the model’s strategy updates too quickly, it tends to assign such samples prematurely to the long-chain mode, degenerating the algorithm to a simple binary classifier based on initial accuracy rather than a collaborative learning of mode selection and accuracy improvement.

\subsection{Details of Warm-up Distillation} 

\begin{table*}[t]
\centering
\small
\resizebox{\textwidth}{!}{
\begin{tabular}{l l c r c r c r c r}
\toprule
\multirow{2}{*}{\bf Model} & \bf \multirow{2}{*}{\bf Mode \& Teacher} & \multicolumn{2}{c}{\textbf{AIME 2024}} & \multicolumn{2}{c}{\textbf{Minerva Algebra}} & \multicolumn{2}{c}{\textbf{Math-500}} & \multicolumn{2}{c}{\textbf{GSM8K}}\\
\cmidrule(lr){3-4} \cmidrule(lr){5-6} \cmidrule(lr){7-8}  \cmidrule(lr){9-10}
& & Pass@1 & \#Tokens & Pass@1 & \#Tokens & Pass@1 & \#Tokens & Pass@1 & \#Tokens \\

\midrule
\textbf{Qwen2.5-1.5B-Instruct} & Short (Base) & 0.0200 & 1300 &  0.7771 & 933 & 0.5168 & 855 & 0.7022 & 466 \\
\textbf{Qwen2.5-Math-1.5B-Instruct} & Short (Base)  & 0.1133 & 1128 & 0.9184 & 586 & 0.7604 & 721 & 0.8572 & 447 \\
\textbf{DeepSeek-R1-1.5B}  &  Long (Base)  & 0.2800 &  18063 & 0.9577 & 3029 & 0.8608 & 5675 & 0.8347 & 1919 \\

\midrule
 \multirow{2}{*}{\bf R1-1.5B + OpenR1-97K} &  \cellcolor{gray!25}$\blacklozenge$ \cellcolor{gray!25} Long (R1-671B) & \cellcolor{gray!25} 0.2933	& \cellcolor{gray!25} 18880	 & \cellcolor{gray!25} 0.9456 & \cellcolor{gray!25} 3239	& \cellcolor{gray!25} 0.8336	& \cellcolor{gray!25} 6780 & \bf \cellcolor{gray!25} 0.8382 & \cellcolor{gray!25} 3423 \\
& $\lozenge$ Short (QMath-1.5B)  & 0.0733	& 3023	& 0.9001 & 642 & 0.7212	& 1004 & 0.7985 & 433 \\

\midrule
\multirow{2}{*}{\bf R1-1.5B + OpenThoughts-114K}  & \cellcolor{gray!25}$\blacklozenge$ Long (R1-671B) & \cellcolor{gray!25} 0.2600 & \cellcolor{gray!25} 17964 & \cellcolor{gray!25} 0.9506 & \cellcolor{gray!25} 3122 & \cellcolor{gray!25} 0.8280 & \cellcolor{gray!25} 6430 & \cellcolor{gray!25} 0.8039 & \cellcolor{gray!25} 2427 \\
 & $\lozenge$ Short (QMath-1.5B) & 0.0800 & 3807 & 0.8947 & 702 & 0.7136 & 1004 & 0.7886 & 449 \\

\midrule
\multirow{2}{*}{\bf R1-1.5B + OpenThoughts-1M}  & \cellcolor{gray!25}$\blacklozenge$ Long (R1-671B) & \bf \cellcolor{gray!25} 0.3067 & \cellcolor{gray!25} 18685 & \cellcolor{gray!25} \bf 0.9530 & \cellcolor{gray!25} 3064 & \cellcolor{gray!25} \bf 0.8360 &  \cellcolor{gray!25} 6209 & \cellcolor{gray!25} 0.8253 & \cellcolor{gray!25} 2363  \\
 & $\lozenge$ Short (QMath-1.5B) & \bf 0.0800 & 3977 & \bf 0.9051 & 690 & \bf 0.7276 & 989 & \bf 0.8202 & 441 \\
 
\bottomrule
\end{tabular}
}
\caption{The effectiveness of different SFT datasets during the warm-up stage. Since these models have not yet been optimized via reinforcement learning, the control tokens \thinktoken and \emph{<short>} are manually inserted to elicit the desired response patterns.} \label{tbl:warm_up}
\end{table*}

\findingbox{Reasoning LLMs can be a good short response learner.}
In this work, knowledge distillation is deployed for warm-up, serving as a critical step to equip the model with the basic ability to generate both long chains and short responses. In this section, we provide additional implementation details. Specifically, we consider multiple datasets for distillation. We compare three datasets of increasing scale and domain coverage: (1) OpenR1, a mathematics-only dataset with rigorously verified solutions; (2) OpenThoughts-114K, a compact yet multi-domain dataset labeled by DeepSeek-R1-67B, covering mathematics, science, and programming; and (3) OpenThoughts-1M, a large-scale and diverse collection that subsumes the former two. Table~\ref{tbl:warm_up} presents the training results on these datasets. Notably, we find that generating short responses using a long-chain reasoning model is relatively straightforward; even with the smallest dataset, OpenR1-97K, the target model successfully learns to produce short outputs. However, this distillation process may incur some performance degradation. For instance, on the Math-500 benchmark, the accuracy of the distilled hybrid model is slightly lower than that of the original model, which may affect the downstream performance during the RL phase. For the distillation stage, larger and more comprehensive datasets can lead to improved performance. However, the marginal gains diminish as the dataset size increases. For example, expanding the dataset from 114K to 1M results in only a 1\% improvement in long-chain accuracy on the Math-500 benchmark. This work provides a preliminary validation of the effectiveness of simple distillation, and we leave the construction of stronger initial hybrid models as an important direction for future research.

\subsection{Case Study} 

\begin{figure}
    \centering
    \includegraphics[width=\linewidth]{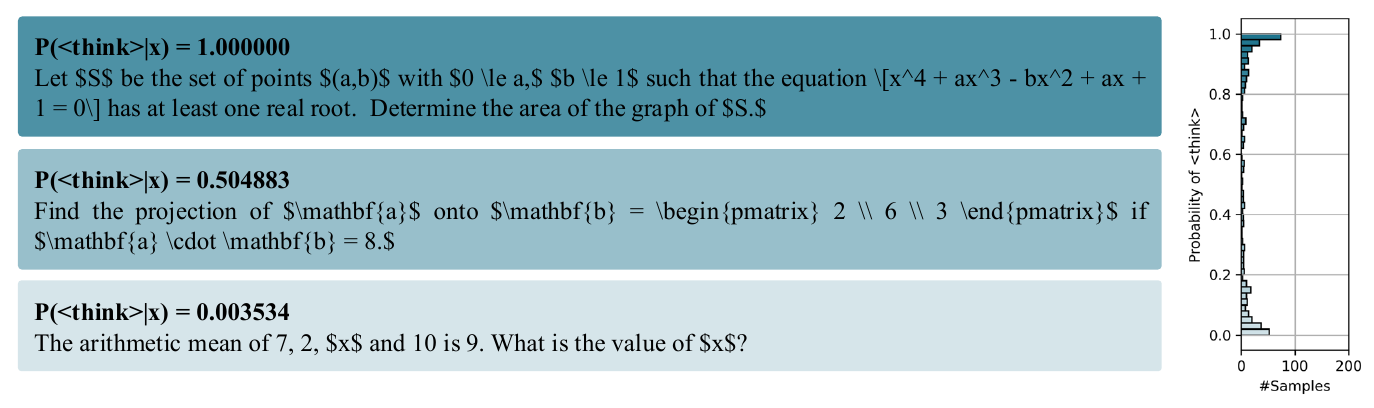}
    \caption{Distribution of the model’s probability of emitting \emph{<think>} on MATH-500. The samples with the highest, medium, and lowest probabilities are highlighted. The example with almost 0 thinking score mainly involves straightforward computation, and the query with 1.0 probability relies more on understanding and logical reasoning. More examples and LLM responses can be found in the appendix.} 
    \label{fig:math500_strategy}
\end{figure}

Figure~\ref{fig:math500_strategy} presents a case study on the model’s predicted probability of selecting the \texttt{<think>} token across the MATH-500 dataset. The distribution reveals that the model makes smooth and hierarchical predictions for queries of different difficulty. In addition, we highlight representative examples corresponding to high, medium, and low confidence levels to illustrate the model’s decision behavior. It can be observed that samples assigned to the short reasoning mode are typically simple arithmetic problems that do not require deep or complex reasoning. In contrast, questions routed to the thinking mode tend to be more complex, involving multiple conditions and concepts. Overall, the results reflect a well-calibrated policy that adapts reasoning depth based on task complexity.

\section{Limitations and Future Works}
This work presents an effective reinforcement learning framework that enables a hybrid model to adapt its inference mode based on both problem complexity and its own capabilities. However, several limitations remain. For instance, during the warm-up phase, we only validate a simple supervised fine-tuning (SFT) approach without extensive parameter tuning to achieve optimal performance, which results in a slight performance drop in the initial model for reinforcement learning. Exploring better strategies for constructing the hybrid model, such as merging techniques or lightweight fine-tuning methods like LoRA to mitigate catastrophic forgetting, could further enhance the overall performance. In addition, while our algorithm has been validated on DeepScaleR, a dataset containing 40K mathematical problems, future work could expand to a broader range of datasets, incorporating more diverse domains to enable more general and practical hybrid reasoning capabilities.

\section{Conclusion}

This paper proposes a reinforcement learning framework for building a hybrid reasoning model. It autonomously decides whether to generate a short response or engage in long-form reasoning based on the complexity of the input. The core of our approach is a Decoupled GRPO algorithm, which separates the reinforcement learning objective into two components: mode selection on the control token and accuracy improvement on the response tokens. This decoupling enables a more balanced contribution between the two learning objectives. Our method effectively reduces unnecessary long-form reasoning, thereby lowering overall system cost and improving user latency.

\clearpage
{
\small
\bibliographystyle{plain}
\bibliography{citation}
}

\end{document}